# Comparative Study and Framework for Automated Summariser Evaluation: LangChain and Hybrid Algorithms

Bagiya Lakshmi S (bagiyalakshmi59@gmail.com), Sanjjushri Varshini R (sanjjushrivarshini@gmail.com), Rohith Mahadevan (rohithmahadev30@gmail.com), Raja CSP Raman (raja.csp@gmail.com)

**Abstract**

Automated Essay Score (AES) is proven to be one of the cutting-edge technologies. Scoring techniques are used for various purposes. Reliable scores are calculated based on influential variables. Such variables can be computed by different methods based on the domain. The research is concentrated on the user's understanding of a given topic. The analysis is based on a scoring index by using Large Language Models. The user can then compare and contrast the understanding of a topic that they recently learned. The results are then contributed towards learning analytics and progression is made for enhancing the learning ability. In this research, the focus is on summarizing a PDF document and gauging a user's understanding of its content. The process involves utilizing a Langchain tool to summarize the PDF and extract the essential information. By employing this technique, the research aims to determine how well the user comprehends the summarized content.

**Keywords:** Learning Analytics, Large Language Models (LLM), LangChain, Summarizing, Automated Essay Score (AES)

## 1 Introduction

The emergence of Automated Essay Scoring (AES) represents a paradigm shift in the realm of educational assessment, heralding cutting-edge technologies that transcend traditional grading methodologies. AES leverages a diverse array of scoring techniques to address multifaceted purposes, ranging from academic evaluation to personalized learning experiences.

At the objective of this research lies a profound investigation into the user's understanding of recently acquired knowledge a quest that transcends conventional evaluation methodologies. Unlike conventional AES, which predominantly targets holistic essay scoring, this research delves into a more specific and granular domain, the evaluation of argumentative essays. However, its ambitions extend beyond mere assessment, it aspires to empower users with the ability to deconstruct and analyze their comprehension of intricate subject matter, offering insights into the fabric of their learning journey. The research capitalizes on a carefully crafted scoring index, a product of Large Language Models, which serves as a beacon of enlightenment, guiding users in comparing and contrasting their understanding of topics recently assimilated into their intellectual repertoire.

It endeavours to bridge the chasm between knowledge acquisition and comprehension through a meticulously structured methodology. It capitalizes on the capabilities of the Langchain tool, an instrument renowned for its prowess in summarizing complex PDF documents while meticulously extracting essential information. This tool becomes the linchpin in a process aimed at gauging the depth of a user's understanding of the summarized content. To this end, metrics of paramount importance employed are cosine similarity, Sorensen Similarity, Jaccard Similarity, and BERT Embedding Similarity. This mathematical measure serves as the arbiter of similarity between the user's interpretation of the PDF content and the compact summary generated by Langchain. It quantifies the congruence between these two sources of information, affording users a tangible score that reflects their grasp of the topic. Furthermore, the research delves even deeper by comparing the user's understanding with the original PDF content, affording yet another score that quantifies the extent of their comprehension. In culmination, these two scores converge to produce a single, definitive value—a percentage that acts as a beacon, shining light on the user's understanding of the PDF topic. A higher percentage signifies a superior level of comprehension, while a lower one points the way to potential areas of improvement, illuminating the path to enriched learning and enhanced educational experiences.

Furthermore, the integration with the Similarity Based Developer Analyzer research [11] promises to enhance candidate selection by incorporating advanced summarization and evaluation techniques. This integration holds the potential to refine the recruitment process, ensuring candidates align not only with job requirements but also with the organization's culture and values.

This research marks a significant step forward, uniting technology, linguistic analysis, and educational empowerment. It leverages the prowess of Large Language Models and advanced scoring techniques to unlock the hidden dimensions of comprehension and elevate the user's learning journey to new heights.

## 2 Literature Review

The field of Automated Essay Scoring (AES) has a rich history spanning over 50 years, drawing substantial interest within the NLP community. Existing AES tools primarily focus on grammar, spelling, and organization but often overlook the critical aspect of persuasive writing. Their paper presents a transformer-based model that achieves above-human accuracy in assessing persuasive discourse elements. Future research will explore model explainability, aiming to facilitate collaborative feedback between teachers and automated systems in the educational context [2]. Their paper applies a financial risk management scoring method to assess university students' study skills and learning styles using computer-based methods, resulting in a single score. It reveals significant differences in study skill scores among students with various learning styles, emphasizing the potential of unsupervised machine learning for understanding and enhancing educational strategies [1]. Their study addresses the crucial aspect of coherence in text quality assessment by introducing a novel approach beyond single-document coherence modelling. Existing methods often overlook the inter-document relationships, which this research seeks to rectify using a Graph Convolutional Network (GCN)-based model.

This approach captures structural similarities between documents by constructing a heterogeneous graph connecting documents based on shared subgraphs. The evaluation of this method on discourse coherence and automated essay scoring tasks demonstrates its superior performance, establishing a new state-of-the-art in both domains [3]. Their study addresses the need for scalable automated essay scoring (AES) in online education by investigating three active learning methods. These approaches, including uncertainty-based, topological-based, and hybrid methods, were evaluated for essay selection and classification using a scoring model trained on transformer-based language representations.

The topological-based method emerged as the most efficient, offering promise for enhancing AES efficiency in online learning environments [4]. The problem of Automated Essay Scoring (AES) has been exacerbated by the proliferation of online learning platforms like Coursera, Udemy, and Khan Academy. Many techniques for AES have been proposed, but they are often constrained by hand-crafted features. In their study, a novel architecture combining Recurrent Neural Networks (RNN) and Convolutional Neural Networks (CNN) was introduced, demonstrating significantly improved grading accuracy compared to other AES systems [5].

In the realm of automated evaluation for student argumentative writing, a survey and organisation of research works have been undertaken, addressing a previously understudied area. Unlike conventional automated writing evaluation, which concentrates on holistic essay scoring, this niche field emphasizes assessing argumentative essays, offering specific feedback encompassing argumentation structures and argument strength trait scores. This meticulous evaluation proves invaluable in enhancing students' critical argumentation skills. Furthermore, the existing literature is structured around tasks, data, and methodologies, and experiments with BERT on representative datasets are conducted to furnish current baselines for this domain [6]. In the domain of automated essay scoring (AES), the utilization of machine learning, particularly natural language processing, has attracted significant attention for its ability to alleviate manual scoring burdens and provide prompt feedback to learners. This study investigates the comparative performance of transformer-based and traditional bag of words (BOW) approaches in AES, employing a dataset of 2088 email responses labelled for politeness. The analysis underscores the superiority of transformer models, even without hyperparameter tuning, over BOW-based logistic regression models, particularly in tasks like politeness classification, where word order and stemming play pivotal roles, ultimately enhancing human rater accuracy [7].

Recent advances in Automated Essay Scoring (AES) have leveraged deep learning techniques, particularly neural networks, to achieve state-of-the-art solutions. While many AES systems assign overall scores to essays, there's a growing need to assess essays based on specific traits and provide tailored feedback to learners. This paper presents a framework that enhances the validity and accuracy of a baseline neural-based AES model, focusing on trait evaluation and scoring. The framework extends the model to predict essay traits, improving the overall score prediction. Among the deep learning models explored Long Short-Term Memory (LSTM) systems outperformed the baseline by 4.6% in terms of quadratic weighted Kappa (QWK). The extended model is applied in iAssistant, an educational module providing learners with trait-specific adaptive feedback [8]. This study focuses on UKARA, an automatic essay scoring system combining NLP and machine learning, using datasets A and B from the UKARA challenge. Due to dataset size limitations, data augmentation techniques, including Synonym Replacement, Random Insertion, Random Swap, and Random Deletion, were employed. The

BiLSTM method created the model, and evaluated using metrics like accuracy, precision, recall, and F-measure with a confusion matrix. Results indicate dataset A achieved 85.07% accuracy without augmentation, while dataset B, with EDA insertion, scored 72.78% accuracy using k-fold cross-validation [9].

This literature review explores the use of similarity techniques in Automated Essay Scoring (AES) systems over the past decade (2010-2020). While various similarity methods have been developed, only a select few have gained prominence in AES. The review, conducted using the Kitchenham method, encompasses 34 articles that meet specific inclusion and quality criteria. It identifies two main categories of similarity techniques used in AES and diverse approaches for scoring student answers. Notably, AES systems often combine multiple methods for improved performance. Furthermore, the review highlights the frequent use of the quadratic weighted kappa (QWK) as the preferred evaluation metric for AES systems [10].

## 3 Methodology and Component Explanation

Figure 1: Architecture diagram: Automated Summariser Evaluation

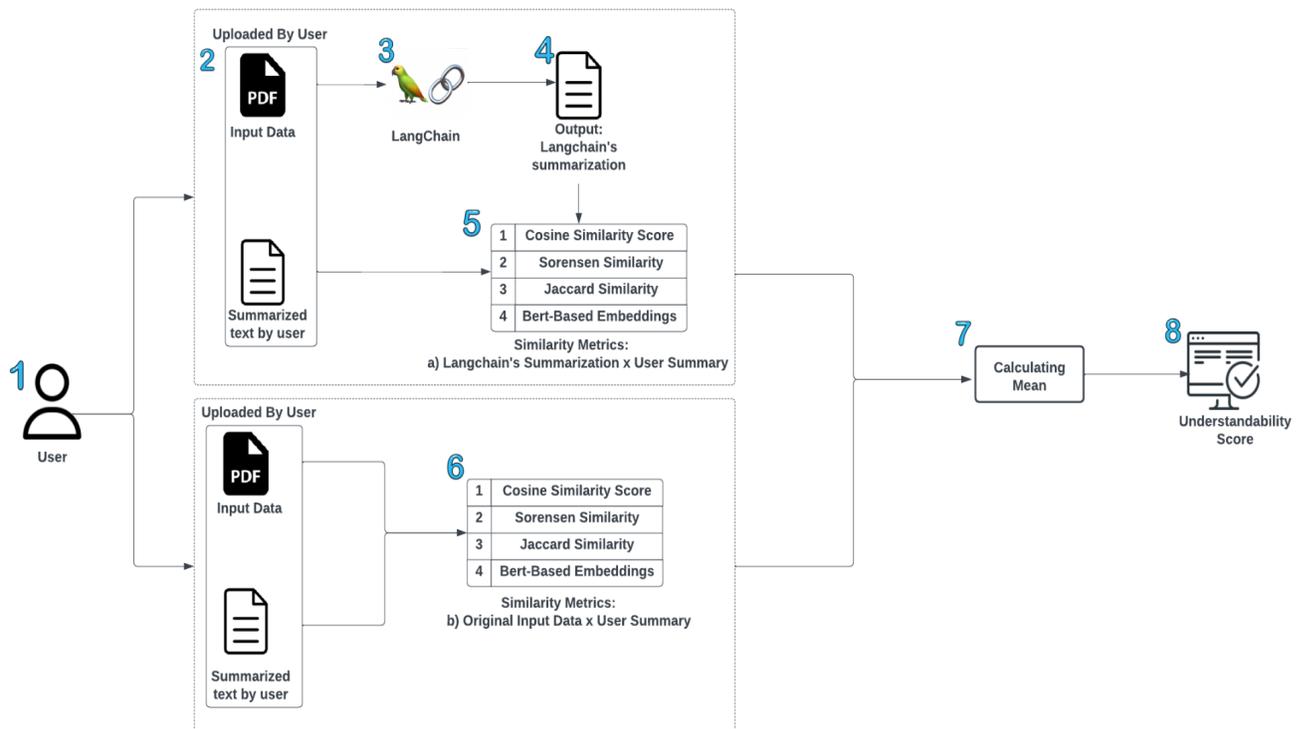

In this section, we elucidate the architectural components and the methodology employed for assessing a user's understanding of PDF content. The architectural diagram above provides a visual representation of our approach.

### 3.1 Evaluation Methodology

Our evaluation methodology hinges primarily on cosine similarity, a mathematical technique employed to gauge the similarity between two sets of data, forming the cornerstone of our approach. Alongside cosine similarity, we also incorporated alternative similarity metrics, including Sorensen Similarity and Jaccard Similarity, recognized for their effectiveness in measuring set-based similarity. Furthermore, we harnessed advanced deep learning techniques to create Bert-based embeddings, enabling us to generate dense vector representations of text, thus enhancing the semantic understanding of our assessment. This comprehensive evaluation framework ensured a thorough assessment of user understanding.

### 3.2 Summarization Using LangChain

To initiate the assessment, we utilize "LangChain," an advanced natural language processing framework designed for summarization tasks. LangChain is employed to summarize PDF content, creating a condensed version of the information that retains key insights and context.

### 3.3 Summarizing Content and User's Understanding Comparison

Following the summarization process using LangChain, we compare this summarized content with the user's understanding of the PDF content. Cosine similarity is applied to this comparison, generating a similarity score that indicates the degree of alignment between the summarized content and the user's comprehension.

### 3.4 Original PDF Content and User's Understanding Comparison

In addition to the above step, we also compare the user's understanding with the original, unaltered content of the PDF. This second comparison, again utilizing cosine similarity, measures how effectively the user comprehends the topic presented in the PDF.

### 3.5 Mean Score Calculation

The results of the two comparisons yield two similarity scores. To provide a comprehensive assessment of the user's comprehension, we calculate the mean (average) of these two scores. This mean score encapsulates the user's level of understanding, accounting for the quality of the summary as well as the alignment between the user's understanding and the original PDF content.

### 3.6 Benchmarking metrics

We conducted benchmarking experiments to explore the effectiveness of various similarity metrics and embedding techniques, including Sorenson, Jaccard, Cosine, and BERT-based embeddings. Our findings demonstrated that BERT-based cosine similarity exhibited superior performance in assessing user understanding, highlighting its capability to capture nuanced semantic relationships within the text.

This architectural diagram and methodology encapsulate our approach to evaluating the user's comprehension of PDF content, combining mathematical similarity metrics with advanced text summarization techniques using "LangChain" for a robust assessment process.

**4 Proposed Solution**

The Bert-based cosine similarity method is used to assess the level of understanding of the user. By using this metric, we are able to compare the summarized content and the user's understanding of the PDF content. The research calculates a score to indicate the degree of similarity between two sources of information in order to measure the degree of similarity between the two.

Furthermore, a comparison is made between the original PDF content and the user's understanding to obtain another score. This comparison allows evaluating the extent to which the user comprehends the topic of the PDF.

It is ultimately the mean of both scores that determines the output of the research. An evaluation of this value can be used to determine the extent to which the user understands the subject matter of the PDF document. A higher percentage indicates a better understanding, while a lower percentage suggests room for improvement.

**4.1 Cosine Similarity Score:**
Cosine similarity is a metric used to measure the similarity between two vectors in a multi-dimensional space. It calculates the cosine of the angle between the vectors, indicating how closely they align. In the context of text analysis, cosine similarity is often employed to compare the similarity of document vectors or word embeddings. A higher cosine similarity score implies greater similarity, with a value of 1 indicating identical vectors and 0 indicating no similarity.

**4.2 Sorensen Similarity:**
Sorensen similarity is a measure of similarity between two sets. It is calculated by dividing the size of the intersection of the sets by the size of their union. In text analysis, it is commonly used to compare the similarity between two sets of words, such as in natural language processing tasks like text classification or clustering. The Sorensen score ranges from 0 (no similarity) to 1 (complete similarity).

**4.3 Jaccard Similarity:**
Jaccard similarity is a specific case of Sorensen similarity and is widely used in set-based comparisons. It is particularly valuable for tasks like document clustering, recommendation systems, and finding common elements between datasets.

**4.4 BERT Embedding Similarity:**
BERT (Bidirectional Encoder Representations from Transformers) is a state-of-the-art pre-trained deep learning model for natural language understanding. BERT embeddings represent words or sentences as dense vectors in a high-dimensional space. The similarity between text fragments is often measured by calculating the cosine similarity between their BERT embeddings. This approach leverages the rich contextual information captured by BERT and is used in various NLP tasks like semantic similarity assessment, sentiment analysis, and document retrieval, where a higher similarity score indicates greater semantic likeness between texts.

Overall, this research employs Langchain to summarise a PDF document and utilizes cosine similarity to assess a user's understanding of the content. By calculating and averaging the similarity scores, the research provides a percentage reflecting the user's comprehension of the PDF topic.

## 5 Results and Analysis

In this section, we present the results of our study, which aimed to assess user understanding of PDF content using a combination of text summarization and similarity metrics. We also provide a comprehensive analysis of the outcomes, shedding light on the performance of different similarity metrics and the impact of text summarization using LangChain.

### 5.1 Performance Metrics

The evaluation was conducted on a diverse set of PDF documents, and performance was measured using accuracy scores. The following table summarizes the results obtained:

Figure 2: Similarity Metrics

| S.NO. | Similarity Metrics | Score |
|---|---|---|
| 1 | Cosine Similarity Score | 63.00% |
| 2 | Sorensen Similarity | 77.04% |
| 3 | Jaccard Similarity | 62.67% |
| 4 | Bert-Based Embeddings | 85.84% |

Our study indicates that Bert-based embeddings, combined with cosine similarity, provide a robust and highly accurate method for assessing user understanding of PDF content. The significant accuracy improvement underscores the importance of utilizing advanced natural language processing techniques in information assessment tasks.

## 6 Future Scope

The future prospects for this research and methodology are exceptionally promising. It has the potential to bring about significant advancements in education and the IT sector. In education, it can reshape the way we evaluate students, offering tailored insights for personalized learning experiences. In the IT industry, it can greatly enhance employee evaluations, ensuring more effective information assimilation and utilization. Moreover, its integration into real-time interactions holds the promise of delivering instant feedback on comprehension, making it a versatile tool with wide-reaching applications.

## 7 Conclusion

In conclusion, our research paper presents a comprehensive methodology for assessing user understanding of PDF content. We leverage advanced natural language processing techniques, including cosine similarity and BERT-based embeddings, to evaluate the alignment between user comprehension and summarized PDF content. Our benchmarking experiments highlight the effectiveness of BERT-based cosine similarity in capturing semantic nuances. This approach offers a robust and nuanced assessment of user comprehension, bridging the gap between document summarization and semantic understanding.